\documentclass[runningheads]{llncs}

\usepackage{graphicx}
\usepackage{comment}
\usepackage{amsmath,amssymb} 
\usepackage{color}

\usepackage{microtype}
\usepackage{fontawesome}
\usepackage{hyperref}
\hypersetup{
    colorlinks=true,
    linkcolor=blue,
    filecolor=magenta,      
    urlcolor=cyan,
}
\usepackage{booktabs} 
\usepackage[ruled]{algorithm2e} 
\usepackage{multicol} 

\newcommand{\data}[1]{\textsc{#1}\xspace}
\newcommand{\ABC}{\data{ABC}}
\newcommand{\famous}{\data{Famous}}
\newcommand{\real}{\data{Real}}
\newcommand{\thingi}{\data{Thingi10k}}
\newcommand{\name}{\textsc{Points2Surf}\xspace}

\newcommand\blfootnote[1]{%
  \begingroup
  \renewcommand\thefootnote{}\footnote{#1}%
  \addtocounter{footnote}{-1}%
  \endgroup
}

\begin{document}
\pagestyle{headings}
\mainmatter
\def\ECCVSubNumber{2965}  

\title{\name \\ {\large Learning Implicit Surfaces from Point Clouds}}

\titlerunning{\name: Learning Implicit Surfaces from Point Cloud Patches}

\author{Philipp Erler\inst{1,4} \and
Paul Guerrero\inst{2} \and
Stefan Ohrhallinger\inst{1,3} \and
Michael Wimmer\inst{1} \and
Niloy J. Mitra\inst{2,4}\index{Mitra, Niloy}}

\authorrunning{Erler et al.}

\institute{
$^1$TU Wien \;\;
$^2$Adobe Research  \;\;
$^3$VRVis  \;\;
$^4$University College London \\ \ \\
\faGlobe~\href{https://www.cg.tuwien.ac.at/research/publications/2020/erler-p2s/}{https://www.cg.tuwien.ac.at/research/publications/2020/erler-p2s/}
}


\maketitle

\blfootnote{Published at ECCV 2020: \url{https://eccv2020.eu/}}


\begin{abstract}

A key step in any scanning-based asset creation workflow is to convert unordered point clouds to a surface. Classical methods (e.g. Poisson reconstruction) start to degrade in the presence of noisy and partial scans. Hence, deep learning based methods have recently been proposed to produce complete surfaces, even from partial scans. However, such data-driven methods struggle to generalize to new shapes with large geometric and topological variations. We present \name, a novel \emph{patch-based} learning framework that produces accurate surfaces directly from raw scans without normals.
Learning a prior over a combination of detailed local patches and coarse global information improves generalization performance and reconstruction accuracy.
Our extensive comparison on both synthetic and real data demonstrates a clear advantage of our method over state-of-the-art alternatives on previously unseen classes (on average, \name brings down reconstruction error by 30\% over SPR and by 270\%+ over deep learning based SotA methods) at the cost of longer computation times and a slight increase in small-scale topological noise in some cases. 
Our source code, pre-trained model, and dataset are available at: \url{https://github.com/ErlerPhilipp/points2surf} 

\end{abstract}


\section{Introduction}

\begin{figure}[t]
    \centering
    \includegraphics[width=\textwidth]{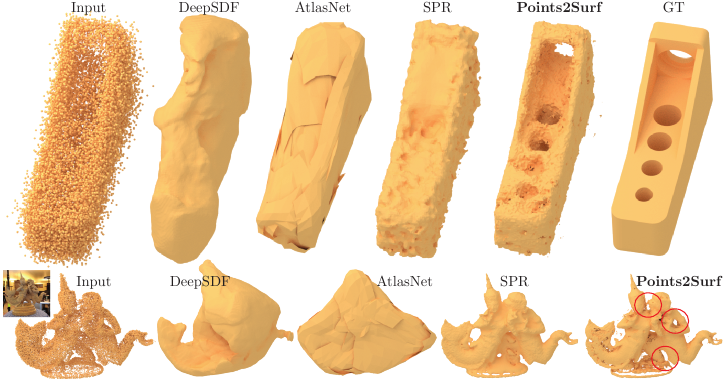}
    \caption{We present \name, a method to reconstruct an accurate implicit surface from a noisy point cloud. Unlike current data-driven surface reconstruction methods like DeepSDF and AtlasNet, it is patch-based, improves detail reconstruction, and unlike Screened Poisson Reconstruction (SPR), a learned prior of low-level patch shapes improves reconstruction accuracy. 
    Note the quality of reconstructions, both geometric and topological, against the original surfaces. The ability of \name to generalize to new shapes makes it the first learning-based approach with significant generalization ability under both geometric and topological variations. 
    }
    \label{fig:teaser}
\end{figure}

Converting unstructured point clouds to surfaces is a key step of most scanning-based asset creation workflows, including games and AR/VR applications. While scanning technologies have become more easily accessible (e.g., depth cameras on smart phones, portable scanners), algorithms for producing a surface mesh remain limited. A good surfacing algorithm should be able to handle raw point clouds with noisy and varying sampling density, work with different surface topologies, and generalize across a large range of scanned shapes. 

Screened Poisson Reconstruction (SPR)~\cite{kazhdan2013screened} is the most commonly used method to convert an unstructured point cloud, along with its per-point normals, to a surface mesh. While the method is  general, in absence of any data-priors, SPR typically produces smooth surfaces, can incorrectly close off holes and tunnels in noisy or non-uniformly sampled scans (see Figure~\ref{fig:teaser}), and further degenerates when per-point normal estimates are erroneous. 

Hence, several data-driven alternatives~\cite{saurer2017point,dai2019scan2mesh,park2019deepsdf,groueix2018atlasnet} have recently been proposed. These methods specialize to particular object categories (e.g., cars, planes, chairs), and typically regress a global latent vector from any input scan. The networks can then decode a final shape (e.g., a collection of primitives~\cite{groueix2018atlasnet} or a mesh~\cite{park2019deepsdf}) from the estimated global latent vector. While such data-specific approaches  handle noisy and partial scans, the methods do \textit{not} generalize to new surfaces with varying shape and topology (see Figure~\ref{fig:teaser}). 

As a solution, we present \name, a method that learns to produce implicit surfaces directly from raw point clouds. During test time, our method can reliably handle raw scans to reproduce fine-scale data features even from noisy and non-uniformly sampled point sets, works for objects with varying topological attributes, and generalizes to new objects (see  Figure~\ref{fig:teaser}). 

Our key insight is to decompose the problem into learning a global and a local function. For the global function, we learn the sign (i.e., inside or outside) of an implicit signed distance function, while, for the local function, we use a patch-based approach to learn absolute distance fields with respect to local point cloud patches. The global task is coarse (i.e., to learn the inside/outside of objects) and hence can be generalized across significant shape variations. The local task exploits the geometric observation that a large variety of shapes can be expressed in terms of a much smaller collection of atomic shape patches~\cite{Badki2020MeshletPF}, which generalizes across arbitrary shapes. 
We demonstrate that such a factorization leads to a simple, robust, and generalizable approach to learn an implicit signed distance field, from which a final surface is extracted using Marching Cubes~\cite{lorensen1987marching}.

We test our algorithms on a range of synthetic and real examples, compare on unseen classes against both classical (reduction in reconstruction error by 30\% over SPR) and learning based strong baselines (reduction in reconstruction error by 470\% over  DeepSDF~\cite{park2019deepsdf} and 270\% over AtlasNet~\cite{groueix2018atlasnet}), and provide ablations studies. We consistently demonstrate both qualitative and quantitative improvement over all the methods that can be applied directly on raw scans.

\section{Related Work}

Several methods have been proposed to reconstruct surfaces from point clouds. We divide these into methods that aggregate information from a large dataset into a data-driven prior, and methods that do not use a data-driven prior.

\paragraph{Non-data-driven surface reconstruction.}
Berger et al.~\cite{berger2017survey} present an in-depth survey that is focused primarily on non-data-driven methods. Here we focus on approaches that are most relevant to our method.
Scale space meshing~\cite{digne2011scale} applies iterative mean curvature motion to smooth the points for meshing.
It preserves multi-resolution features well. Ohrhallinger et al. propose a combinatorial method~\cite{ohrhallinger2013minimizing} which compares favorably with previous methods such as Wrap~\cite{edelsbrunner2003surface}, TightCocone~\cite{dey2003tight} and Shrink~\cite{chaine2003geometric} especially for sparse sampling and thin structures. However, these methods are not designed to process noisy point clouds.
Another line of work deforms initial meshes~\cite{sharf2006competing,li2010arterial} or parametric patches~\cite{williams2019deep} to fit a noisy point cloud. These approaches however, cannot change the topology and connectivity of the original meshes or patches, usually resulting in a different connectivity or topology than the ground truth.
The most widely-used approaches to reconstruct surfaces with arbitrary topology from noisy point clouds fit implicit functions to the point cloud and generate a surface as a level set of the function. Early work by Hoppe et al. introduced this approach~\cite{hoppe1992surface}, and since then several methods have focused on different representations of the implicit functions, like Fourier coefficients~\cite{kazhdan2005reconstruction}, wavelets~\cite{manson2008streaming}, radial-basis functions~\cite{ohtake20053d} or multi-scale approaches~\cite{ohtake2003multi,nagai2009smoothing}. Alliez et al.~\cite{alliez2007voronoi} use a PCA of 3D Voronoi cells to estimate gradients and fit an implicit function by solving an eigenvalue problem. This approach tends to over-smooth geometric detail. Poisson reconstruction~\cite{kazhdan2006poisson,kazhdan2013screened} is the current gold standard for non-data-driven surface reconstruction from point clouds. None of the above methods make use of a prior that distills information about about typical surface shapes from a large dataset. Hence, while they are very general, they fail to handle partial and/or noisy input. We provide extensive comparisons to Screened Poisson Reconstruction (SPR)~\cite{kazhdan2013screened} in Section~\ref{sec:results}.

\paragraph{Data-driven surface reconstruction.}
Recently, several methods have been proposed to learn a prior of typical surface shapes from a large dataset.
Early work was done by Sauerer et al.~\cite{saurer2017point}, where a  decision tree is trained to predict the absolute distance part of an SDF, but ground truth normals are still required to obtain the sign (inside/outside) of the SDF.
More recent data-driven methods represent surfaces with a single latent feature vector in a learned feature space. An encoder can be trained to obtain the feature vector from a point cloud. The feature representation acts as a strong prior, since only shapes that are representable in the feature space are reconstructed.
Early methods use voxel-based representations of the surfaces, with spatial data-structures like octrees offsetting the cost of a full volumetric grid~\cite{tatarchenko2017octree,wang2018adaptive}.
Scan2Mesh~\cite{dai2019scan2mesh} reconstructs a coarse mesh, including vertices and triangles, from a scan with impressive robustness to missing parts. However, the result is typically very coarse and not watertight or manifold, and does not apply to arbitrary new shapes.
AtlasNet~\cite{groueix2018atlasnet} uses multiple parametric surfaces as representation that jointly form a surface, achieving impressive accuracy and cross-category generalization.
More recently, several approaches learn implicit function representations of surfaces~\cite{park2019deepsdf,Chen:2019:implicit,Mescheder:2019:Occupancy}. These methods are trained to learn a functional that maps a latent encoding of a surface to an implicit function that can be used to extract a continuous surface. The implicit representation is more suitable for surfaces with complex topology and tends to produce aesthetically pleasing smooth surfaces. 

The single latent feature vector that the methods above use to represent a surface acts as a strong prior, allowing these methods to reconstruct surfaces even in the presence of strong noise or missing parts; but it also limits the generality of these methods. The feature space typically captures only shapes that are similar to the shapes in the training set, and the variety of shapes that can be captured by the feature space is limited by the fixed capacity of the latent feature vector. Instead, we propose to decompose the SDF that is used to reconstruct the surface into a coarse global sign and a detailed local absolute distance. Separate feature vectors are used to represent the global and local parts, allowing us to represent detailed local geometry, without losing coarse global information about the shape.

\section{Method}

Our goal is to reconstruct a watertight surface $S$ from a 3D point cloud $P = \{p_1, ..., p_N\}$ that was sampled from the surface $S$ through a noisy sampling process, like a 3D scanner. We represent a surface as the zero-set of a Signed Distance Function (SDF) $f_S$:
\begin{equation}
    S = L_0(f_S) = \{x \in \mathbb{R}^3\ |\ f_S(x)=0\}.
\end{equation}
Recent work~\cite{park2019deepsdf,Chen:2019:implicit,Mescheder:2019:Occupancy} has shown that such an implicit representation of the surface is particularly suitable for neural networks, which can be trained as functionals that take as input a latent representation of the point cloud and output an approximation of the SDF:
\begin{equation}
\label{eq:global_z}
    f_S(x) \approx \tilde{f}_P(x) = s_{\theta}(x | z),\ \text{with}\ z = e_{\phi}(P),
\end{equation}
where $z$ is a latent description of surface $S$ that can be encoded from the input point cloud with an encoder $e$, and $s$ is implemented by a neural network that is conditioned on the latent vector $z$. The networks $s$ and $e$ are parameterized by $\theta$ and $\phi$, respectively. This representation of the surface is continuous, usually produces watertight meshes, and can naturally encode arbitrary topology. Different from non-data-driven methods like SPR~\cite{kazhdan2013screened}, the trained network obtains a strong prior from the dataset it was trained on, that allows robust reconstruction of surfaces even from unreliable input, such as noisy and sparsely sampled point clouds. However, encoding the entire surface with a single latent vector imposes limitations on the accuracy and generality of the reconstruction, due to the limited capacity of the latent representation.

In this work, we factorize the SDF into the absolute distance $f^d$ and the sign of the distance $f^s$, and take a closer look at the information needed to approximate each factor. To estimate the absolute distance $\tilde{f}^d(x)$ at a query point $x$, we only need a \textit{local} neighborhood of the query point:
\begin{equation}
    \tilde{f}_P^d(x) = s^d_{\theta}(x | z^d_x),\ \text{with}\ z^d_x = e^d_{\phi}(\mathbf{p}^d_x),
\end{equation}
where $\mathbf{p}^d_x \subset P$ is a local neighborhood of the point cloud centered around $x$.
Estimating the distance from an encoding of a local neighborhood gives us more accuracy than estimating it from an encoding of the entire shape, since the local encoding $z^d_x$ can more accurately represent the local neighborhood around $x$ than the global encoding $z$. Note that in a point cloud without noise and sufficiently dense sampling, the single closest point $p^* \subset P$ to the query $x$ would be enough to obtain a good approximation of the absolute distance. But since we work with noisy and sparsely sampled point clouds, using a larger local neighborhood increases robustness.

In order to estimate the sign $\tilde{f}^s(x)$ at the query point $x$, we need \textit{global} information about the entire shape, since the interior/exterior of a watertight surface cannot be estimated reliably from a local patch. Instead, we take a global sub-sample of the point cloud $P$ as input:
\begin{equation}
    \tilde{f}_P^s(x) = \mathrm{sgn}\big(\tilde{g}_P^s(x)\big) = \mathrm{sgn}\big(s^s_{\theta}(x | z^s_x)\big),\ \text{with}\ z^s_x = e^s_{\psi}(\mathbf{p}^s_x),
\end{equation}
where $\mathbf{p}^s_x \subset P$ is a global subsample of the point cloud, $\psi$ are the parameters of the encoder, and $\tilde{g}_P^s(x)$ are logits of the probability that $x$ has a positive sign. Working with logits avoids discontinuities near the surface, where the sign changes. Since it is more important to have accurate information closer to the query point, we sample $\mathbf{p}^s_x$ with a density gradient that is highest near the query point and falls off with distance from the query point.

We found that sharing information between the two latent descriptions $z^s_x$ and $z^d_x$ benefits both the absolute distance and the sign of the SDF, giving us the formulation we use in Points2Surf:
\begin{equation}
    \big(\tilde{f}_P^d(x), \tilde{g}_P^s(x)\big) = s_{\theta}(x | z^d_x, z^s_x),\ \text{with}\ z^d_x = e^d_{\phi}(\mathbf{p}^d_x) \ \text{and}\ z^s_x = e^s_{\psi}(\mathbf{p}^s_x).
\end{equation}
We describe the architecture of our encoders and decoder, the training setup, and our patch sampling strategy in more detail in Section~\ref{sec:architecture}.

To reconstruct the surface S, we apply Marching Cubes~\cite{lorensen1987marching} to a sample grid of the estimated SDF $\tilde{f}^d(x) * \tilde{f}^s(x)$. In Section~\ref{sec:recon}, we describe a strategy to improve performance by only evaluating a subset of the grid samples.

\begin{figure}[t]
    \centering
    \includegraphics[width=\linewidth]{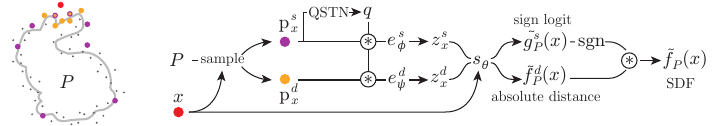}
    \caption{Points2Surf Architecture. Given a query point $x$ (red) and a point cloud $P$ (gray), we sample a local patch (yellow) and a coarse global subsample (purple) of the point cloud. These are encoded into two feature vectors that are fed to a decoder, which outputs a logit of the sign probability and the absolute distance of the SDF at the query point $x$.}
    \label{fig:architecture}
\end{figure}

\subsection{Architecture and Training}
\label{sec:architecture}

Figure~\ref{fig:architecture} shows an overview of our architecture.
Our approach estimates the absolute distance $\tilde{f}_P^d(x)$ and the sign logits $\tilde{g}_P^s(x)$ at a query point based on two inputs: the query point $x$ and the point cloud $P$.

\paragraph{Pointset sampling.}
The local patch $\mathbf{p}^d_x$ and the global sub-sample $\mathbf{p}^s_x$ are both chosen from the point cloud $P$ based on the query point $x$. The set  $\mathbf{p}^d_x$ is made of the $n_d$ nearest neighbors of the query point (we choose $n_d = 300$ but also experiment with other values). Unlike a fixed radius, the nearest neighbors are suitable for query points with arbitrary distance from the point cloud. The global sub-sample $\mathbf{p}^s_x$ is sampled from $P$ with a density gradient that decreases with distance from $x$:
\begin{equation}
    \rho(p_i) = \frac{v(p_i)}{\sum_{p_j \in P} v(p_j)},\ \text{with}\ v(p_i) = \left[ 1 - 1.5 \frac{\|p_i-x\|_2}{\max_{p_j \in P} \|p_j-x\|_2} \right]_{0.05}^1,
\label{eq:global_subsample}
\end{equation}
where $\rho$ is the sample probability for a point $p_i \in P$, $v$ is the gradient that decreases with distance from $x$, and the square brackets denote clamping. The minimum value for the clamping ensures that some far points are taken and the sub-sample can represent a closed object. We sample $n_s$ points from $P$ according to this probability (we choose $n_s = 1000$ in our experiments).

\paragraph{Pointset normalization.}
Both $\mathbf{p}^d_x$ and $\mathbf{p}^s_x$ are normalized by centering them at the query point, and scaling them to unit radius. After running the network, the estimated distance is scaled back to the original size before comparing to the ground truth. Due to the centering, the query point is always at the origin of the normalized coordinate frame and does not need to be passed explicitly to the network. To normalize the orientation of both point subsets, we use a data-driven approach: a Quaternion Spatial Transformer Network (QSTN)~\cite{GuerreroEtAl:PCPNet:EG:2018} takes as input the global subset $\mathbf{p}^s_x$ and estimates a rotation represented as quaternion $q$ that is used to rotate both point subsets. We take the global subset as input, since the global information can help the network with finding a more consistent rotation. The QSTN is trained end-to-end with the full architecture, without direct supervision for the rotation.

\paragraph{Encoder and decoder architecture.}
The local encoder $e^d_{\phi}$, and the global encoder $e^s_{\psi}$ are both implemented as PointNets~\cite{qi2016pointnet}, sharing the same architecture, but not the parameters. Following the PointNet architecture, a feature representation for each point is computed using $5$ MLP layers, with a spatial transformer in feature space after the third layer. Each layer except the last one uses batch normalization and ReLU. The point feature representations are then aggregated into point set feature representations $z^d_x = e^d_{\phi}(\mathbf{p}^d_x)$ and $z^s_x = e^s_{\psi}(\mathbf{p}^s_x)$ using a channel-wise maximum.
The decoder $s_{\theta}$ is implemented as 4-layer MLP that takes as input the concatenated feature vectors $z^d_x$ and $z^s_x$ and outputs both the absolute distance $\tilde{f}^d(x)$ and sign logits $\tilde{g}^s(x)$.

\paragraph{Losses and training.} We train our networks end-to-end to regress the absolute distance of the query point $x$ from the watertight ground-truth surface $S$ and classify the sign as positive (outside $S$) or negative (inside $S$). We assume that ground-truth surfaces are available during training for supervision. We use an $L_2$-based regression for the absolute distance:
\begin{equation}
    \mathcal{L}^d(x, P, S) = \|\tanh(|\tilde{f}_P^d(x)|) - \tanh(|d(x, S)|)\|_2^2,
\end{equation}
where $d(x, S)$ is the distance of $x$ to the ground-truth surface $S$. The $\tanh$ function gives more weight to smaller absolute distances, which are more important for an accurate surface reconstruction.
For the sign classification, we use the binary cross entropy $H$ as loss:
\begin{equation}
    \mathcal{L}^s(x, P, S) = H\Big(\sigma\big(\tilde{g}_P^s(x)\big),\ [f_S(x) > 0]\Big),
\end{equation}
where $\sigma$ is the logistic function to convert the sign logits to probabilities, and $[f_S(x) > 0]$ is $1$ if $x$ is outside the surface $S$ and $0$ otherwise.
In our optimization, we minimize these two losses for all shapes and query points in the training set:
\begin{equation}
    \sum_{(P, S) \in \mathcal{S}} \sum_{x \in \mathcal{X}_S} \mathcal{L}^d(x, P, S) + \mathcal{L}^s(x, P, S),
\end{equation}
where $\mathcal{S}$ is the set of surfaces $S$ and corresponding point clouds $P$ in the training set and $\mathcal{X}_S$ the set of query points for shape $S$.
Estimating the sign as a classification task instead of regressing the signed distance allows the network to express confidence through the magnitude of the sign logits, improving performance.

\subsection{Surface Reconstruction}
\label{sec:recon}

At inference time, we want to reconstruct a surface $\tilde{S}$ from an estimated SDF $\tilde{f}(x) = \tilde{f}^d(x) * \tilde{f}^s(x)$. A straight-forward approach is to apply Marching Cubes~\cite{lorensen1987marching} to a volumetric grid of SDF samples. Obtaining a high-resolution result, however, would require evaluating a prohibitive number of samples for each shape. We observe that in order to reconstruct a surface, a Truncated Signed Distance Field (TSDF) is sufficient, where the SDF is truncated to the interval $[-\epsilon, \epsilon]$ (we set $\epsilon$ to three times the grid spacing in all our experiments). We only need to evaluate the SDF for samples that are inside this interval, while samples outside the interval merely need the correct sign. We leave samples with a distance larger than $\epsilon$ to the nearest point in $P$ blank, and in a second step, we propagate the signed distance values from non-blank to blank samples, to obtain the correct sign in the truncated regions of the TSDF. We iteratively apply a box filter of size $3^3$ at the blank samples until convergence. 
In each step, we update initially unknown samples only if the filter response is greater than a user-defined confidence threshold (we use 13 in our experiments). After each step, the samples are set to -1 if the filter response was negative or to +1 for a positive response.

\section{Results}
\label{sec:results}

\begin{figure}[t!]
    \centering
    \includegraphics[width=\linewidth]{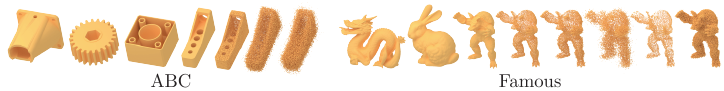}
    \caption{Dataset examples. Examples of the ABC dataset and its three variants are shown on the left, examples of the famous dataset and its five variants on the right.}
    \label{fig:dataset_examples}
\end{figure}

\begin{table}[t]

\caption{Comparison of reconstruction errors. We show the Chamfer distance between reconstructed and ground-truth surfaces averaged over all shapes in a dataset. Both the absolute value of the error multiplied by $100$ (abs.), and the error relative to Point2Surf (rel.) are shown to facilitate the comparison. Our method consistently performs better than the baselines, due to its strong and generalizable prior.}

\centering 
\small
\begin{tabular}{@{}rrrrrrrrrrcc@{}} 
& \multicolumn{2}{c}{DeepSDF} & \phantom{.} & \multicolumn{2}{c}{AtlasNet} & \phantom{.} & \multicolumn{2}{c}{SPR} & \phantom{.} & \multicolumn{2}{c}{\name} \\
\cmidrule{2-3} \cmidrule{5-6} \cmidrule{8-9} \cmidrule{11-12}
& abs. & rel. && abs. & rel. && abs. & rel. && abs. & rel. \\
\toprule 
\ABC no-noise & 8.41 & 4.68 && 4.69 & 2.61 && 2.49 & 1.39 && \textbf{1.80} & \textbf{1.00} \\
\ABC var-noise & 12.51 & 5.86 && 4.04 & 1.89 && 3.29 & 1.54 && \textbf{2.14} & \textbf{1.00} \\
\ABC max-noise & 11.34 & 4.11 && 4.47 & 1.62 && 3.89 & 1.41 && \textbf{2.76} & \textbf{1.00} \\
\famous no-noise & 10.08 & 7.14 && 4.69 & 3.33 && 1.67 & 1.18 && \textbf{1.41} & \textbf{1.00} \\
\famous med-noise & 9.89 & 6.57 && 4.54 & 3.01 && 1.80 & 1.20 && \textbf{1.51} & \textbf{1.00} \\
\famous max-noise & 13.17 & 5.23 && 4.14 & 1.64 && 3.41 & 1.35 && \textbf{2.52} & \textbf{1.00} \\
\famous sparse & 10.41 & 5.41 && 4.91 & 2.55 && 2.17 & 1.12 && \textbf{1.93} & \textbf{1.00} \\
\famous dense & 9.49 & 7.15 && 4.35 & 3.28 && 1.60 & 1.21 && \textbf{1.33} & \textbf{1.00} \\
\midrule
\textbf{average} & 10.66 & 5.77 && 4.48 & 2.49 && 2.54 & 1.30 && \textbf{1.92} & \textbf{1.00}  
\end{tabular}

\label{tab:quant_comparison}
\end{table}

\begin{table}[t]

\caption{Quantitative comparison of reconstruction errors on the Thingi10k dataset. Note that none of the methods was retrained on the Thingi10k in order to test generalization to new data.}

\centering 
\small
\begin{tabular}{@{}rrrrrrrrrrcc@{}} 
& \multicolumn{2}{c}{DeepSDF} & \phantom{.} & \multicolumn{2}{c}{AtlasNet} & \phantom{.} & \multicolumn{2}{c}{SPR} & \phantom{.} & \multicolumn{2}{c}{\name} \\
\cmidrule{2-3} \cmidrule{5-6} \cmidrule{8-9} \cmidrule{11-12}
& abs. & rel. && abs. & rel. && abs. & rel. && abs. & rel. \\
\toprule 
\thingi no-noise  & 9.16  & 6.48 && 5.29 & 3.74 && 1.78 & 1.26 && \textbf{1.41} & \textbf{1.00} \\
\thingi med-noise & 8.83  & 5.99 && 5.19 & 3.52 && 1.81 & 1.23 && \textbf{1.47} & \textbf{1.00} \\
\thingi max-noise & 12.28 & 4.68 && 4.90 & 1.87 && 3.23 & 1.23 && \textbf{2.62} & \textbf{1.00} \\
\thingi sparse    & 9.56  & 4.54 && 5.64 & 2.68 && 2.35 & 1.12 && \textbf{2.11} & \textbf{1.00} \\
\thingi dense     & 8.35  & 6.19 && 5.02 & 3.72 && 1.57 & 1.16 && \textbf{1.35} & \textbf{1.00} \\
\midrule
\textbf{average}  & 9.64  & 5.58 && 5.21 & 3.11 && 2.15 & 1.20 && \textbf{1.79} & \textbf{1.00}  
\end{tabular}

\label{tab:quant_comparison_thingi}
\end{table}

\begin{figure}[t]
    \centering
    \includegraphics[width=\linewidth]{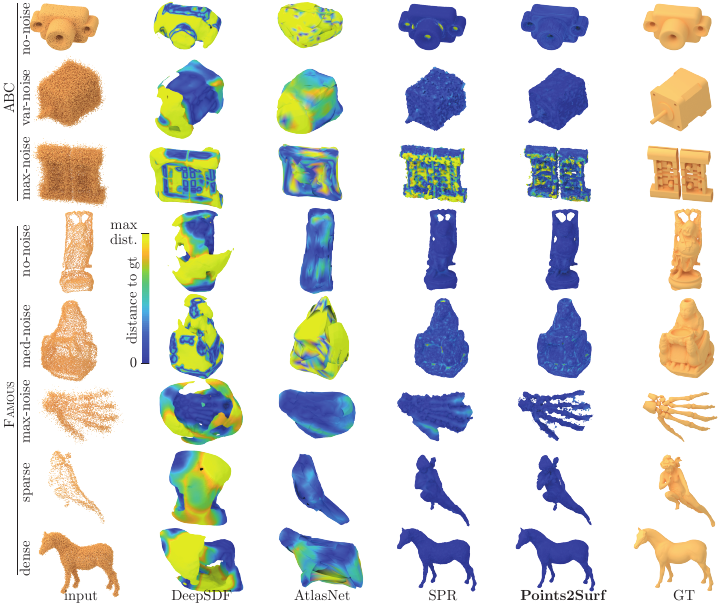}
    \caption{Qualitative comparison of surface reconstructions. We evaluate one example from each dataset variant with each method. Colors show the distance of the reconstructed surface to the ground-truth surface.}
    \label{fig:qual_comparison}
\end{figure}

\begin{figure}[t]
    \centering
    \includegraphics[width=\linewidth]{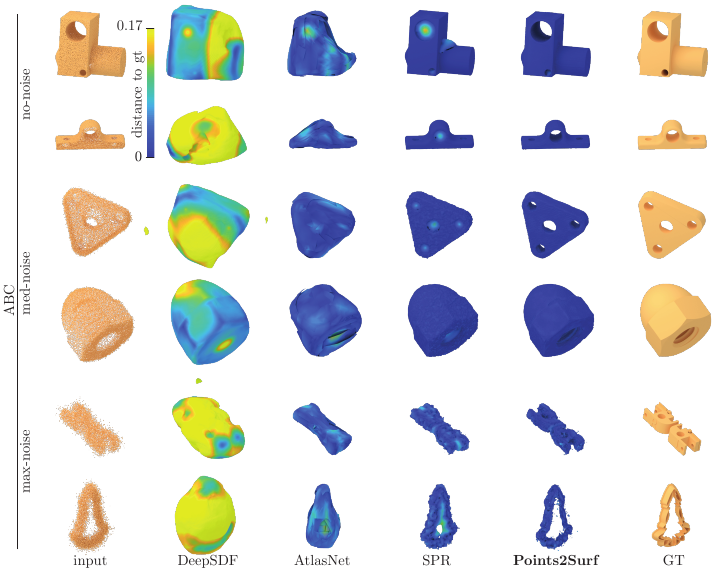}
    \caption{Additional qualitative comparison of surface reconstructions on the \ABC dataset. We evaluate two examples from each dataset variant with each method. Colors show the distance of the reconstructed surface to the ground truth surface.}
    \label{fig:qual_comparison_abc}
\end{figure}

\begin{figure}[p]
    \centering
    \includegraphics[width=\linewidth]{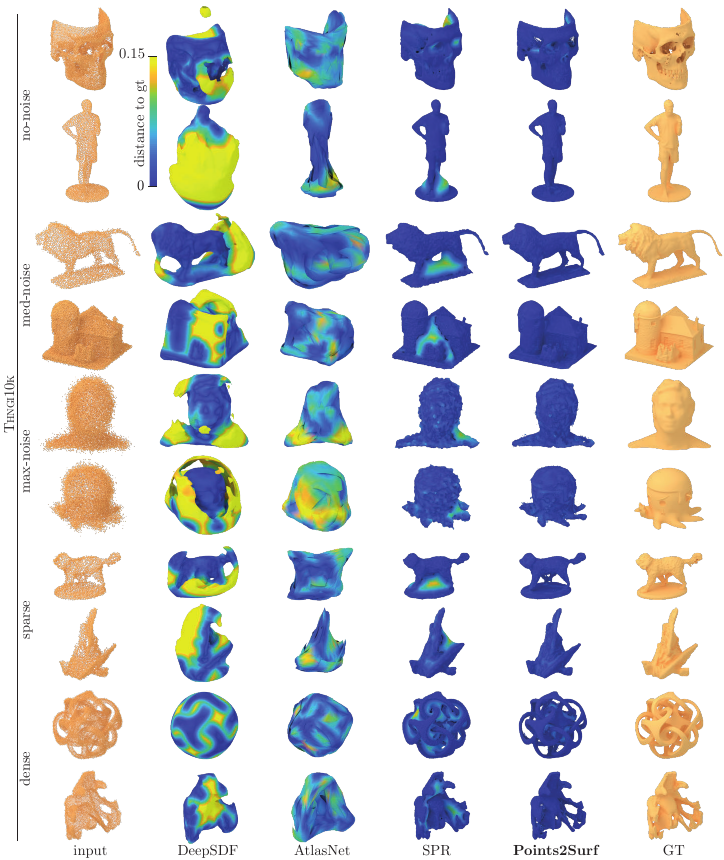}
    \caption{Qualitative comparison of surface reconstructions on the \thingi dataset. We evaluate two examples from each dataset variant with each method. Colors show the distance of the reconstructed surface to the ground truth surface, shades of blue indicating low error.}
    \label{fig:qual_comparison_thingi}
\end{figure}

We compare our method to SPR as the gold standard for non-data-driven surface reconstruction and to two state-of-the-art data-driven surface reconstruction methods. We provide both qualitative and quantitative comparisons on several datasets in Section~\ref{sec:comparison}, perform an ablation study in Section~\ref{sec:ablation}, and provide detailed timings in Section~\ref{sec:timings}

\subsection{Datasets}

We train and evaluate on the \ABC dataset~\cite{Koch_2019_CVPR} which contains a large number and variety of high-quality CAD meshes. We pick $4950$ clean watertight meshes for training and $100$ meshes for the validation and test sets. Note that each mesh produces a large number of diverse patches as training samples.
Operating on local patches also allows us to generalize better,
which we demonstrate on two additional test-only datasets:
a dataset of $22$ diverse meshes which are well-known in geometry processing, such as the Utah teapot and the Stanford Bunny, which we call the \famous dataset, and $3$ \real scans of complex objects used in several denoising papers~\cite{wolff2016noiseremoval,rakotosaona2019pointcleannet}.
Examples from each dataset are shown in Figure~\ref{fig:dataset_examples}. The \ABC dataset contains predominantly CAD models of mechanical parts, while the \famous dataset contains more organic shapes, such as characters and animals. Since we train on the \ABC dataset, the \famous dataset serves to test the generalizability of our method versus baselines.

\paragraph{Pointcloud sampling.} As a pre-processing step, we center all meshes at the origin and scale them uniformly to fit within the unit cube. To obtain point clouds $P$ from the meshes $S$ in the datasets, we simulate scanning them with a time-of-flight sensor from random viewpoints using BlenSor~\cite{gschwandtner2011blensor}. BlenSor realistically simulates various types of scanner noise and artifacts such as backfolding, ray reflections, and per-ray noise. We scan each mesh in the \famous dataset with $10$ scans and each mesh in the \ABC dataset with a random number of scans, between $5$ and $30$. For each scan, we place the scanner at a random location on a sphere centered at the origin, with the radius chosen randomly in $U[3L, 5L]$, where $L$ is the largest side of the mesh bounding box. The scanner is oriented to point at a location with small random offset from the origin, between $U[-0.1L, 0.1L]$ along each axis, and rotated randomly around the view direction. Each scan has a resolution of $176 \times 144$, resulting in roughly $25$k points, minus some points missing due to simulated scanning artifacts. The point clouds of multiple scans of a mesh are merged to obtain the final point cloud.

\paragraph{Dataset variants.} We create multiple versions of both the \ABC and \famous datasets, with varying amount of per-ray noise. This Gaussian noise added to the depth values simulates inaccuracies in the depth measurements. 
For both datasets, we create a noise-free version of the point clouds, called \emph{\ABC no-noise} and \emph{\famous no-noise}. Also, we make versions with strong noise (standard deviation is $0.05L$) called \emph{\ABC max-noise} and \emph{\famous max-noise}. Since we need varying noise strength for the training, we create a version of \ABC where the standard deviation is randomly chosen in $U[0, 0.05L]$ (\emph{\ABC var-noise}), and a version with a constant noise strength of $0.01L$ for the test set (\emph{\famous med-noise}).
Additionally we create sparser ($5$ scans) and denser ($30$ scans) point clouds in comparison to the $10$ scans of the other variants of \famous. Both variants have a medium noise strength of $0.01L$.
Additionally, we show a comparison with scanned objects from the \thingi dataset using the same variants as \famous. We take 100 meshes that are tagged with `scan' or `sculpture'. These objects are mostly animals, humans and faces, many of them realistic, some artistic.
The training set uses the \emph{\ABC var-noise} version, all other variants are used for evaluation only.

\paragraph{Query points.} The training set also contains a set $\mathcal{X}_S$ of query points for each (point cloud, mesh) pair $(P, S) \in \mathcal{S}$. Query points closer to the surface are more important for the surface reconstruction and more difficult to estimate. Hence, we randomly sample a set of $1000$ points on the surface and offset them in the normal direction by a uniform random amount in $U[-0.02L, 0.02L]$. An additional $1000$ query points are sampled randomly in the unit cube that contains the surface, for a total of $2000$ query points per mesh. During training, we randomly sample a subset of $1000$ query points per mesh in each epoch.

\subsection{Comparison to Baselines}
\label{sec:comparison}

We compare our method to recent data-driven surface reconstruction methods, AtlasNet~\cite{groueix2018atlasnet} and DeepSDF~\cite{park2019deepsdf}, and to SPR~\cite{kazhdan2013screened}, which is still the gold standard in non-data-driven surface reconstruction from point clouds. Both AtlasNet and DeepSDF represent a full surface as a single latent vector that is decoded into either a set of parametric surface patches in AtlasNet, or an SDF in DeepSDF. In contrast, SPR solves for an SDF that has a given sparse set of point normals as gradients, and takes on values of $0$ at the point locations. We use the default values and training protocols given by the authors for all baselines (more details in the Supplementary) and re-train the two data-driven methods on our training set. We provide SPR with point normals as input, which we estimate from the input point cloud using the recent PCPNet~\cite{GuerreroEtAl:PCPNet:EG:2018}. 

For DeepSDF, we followed the method in the original paper with minor adaptations. For the training set, we take our query points and the corresponding Signed Distances (SD) as GT SDF samples. For the test sets, we take the point clouds from our dataset. For each point, we generate 2 SDF samples, one in positive and one in negative normal direction, with random offset. The normals are taken from the ground truth face closest to the sample. The corresponding SDs are the +- normal offset. We add 20\% random samples from the unit cube with GT SD.

\paragraph{Error metric.} To measure the reconstruction error of each method, we sample both the reconstructed surface and the ground-truth surface with $10$k points and compute the Chamfer distance~\cite{Barrow:1977:Chamfer,fan2017point} between the two point sets:
\begin{equation}
\label{eq:chamfer}
d_{\text{ch}}(A, B) := \frac{1}{|A|} \sum_{p_i \in A} \min_{p_j \in B} \|p_i - p_j\|^2_2\ + \frac{1}{|B|} \sum_{p_j \in B} \min_{p_i \in A} \|p_j - p_i\|^2_2,
\end{equation}
where $A$ and $B$ are point sets sampled on the two surfaces. The Chamfer distance penalizes both false negatives (missing parts) and false positives (excess parts).

\paragraph{Quantitative and qualitative comparison.} A quantitative comparison of the reconstruction quality is shown in Tables~\ref{tab:quant_comparison} and~\ref{tab:quant_comparison_thingi}. Figures~\ref{fig:qual_comparison},~\ref{fig:qual_comparison_abc} and~\ref{fig:qual_comparison_thingi} compare a few reconstructions qualitatively. All methods were trained on the training set of the \emph{\ABC var-noise} dataset, which contains predominantly mechanical parts, while the more organic shapes in the \famous dataset test how well each method can generalize to novel types of shapes.

Both DeepSDF and AtlasNet use a global shape prior, which is well suited for a dataset with high geometric consistency among the shapes, like cars in ShapeNet, but struggles with the significantly larger geometric and topological diversity in our datasets, reflected in a higher reconstruction error than SPR or \name.
This is also clearly visible in Figure~\ref{fig:qual_comparison}, where the surfaces reconstructed by DeepSDF and AtlasNet appear over-smoothed and inaccurate.

In SPR, the full shape space does not need to be encoded into a prior with limited capacity, resulting in a better accuracy. But this lack of a strong prior also prevents SPR from robustly reconstructing typical surface features, such as holes or planar patches (see Figures~\ref{fig:teaser} and \ref{fig:qual_comparison}).

\name learns a prior of local surface details, instead of a prior for global surface shapes. This local prior helps recover surface details like holes and planar patches more robustly, improving our accuracy over SPR. Since there is less variety and more consistency in local surface details compared to global surface shapes, our method generalizes better and achieves a higher accuracy than the data-driven methods that use a prior over the global surface shape.

\paragraph{Generalization.}
A comparison of our generalization performance against AtlasNet and DeepSDF shows an advantage for our method. In Table~\ref{tab:quant_comparison}, we can see that the error for DeepSDF and AtlasNet increases more when going from the \ABC dataset to the \famous dataset than the error for our method.
This suggests that our method generalizes better from the CAD models in the \ABC dataset set to the more organic shapes in the \famous dataset.

\begin{figure}[t!]
    \centering
    \includegraphics[width=\linewidth]{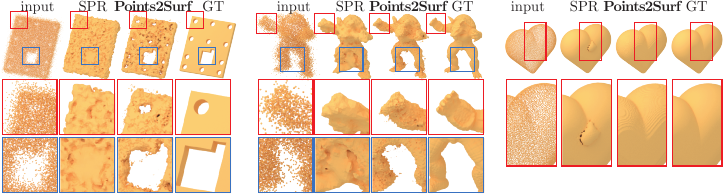}
    \caption{Comparison of reconstruction details. Our learned prior improves the reconstruction robustness for geometric detail compared to SPR.}
    \label{fig:detail_comparison}
\end{figure}

\paragraph{Topological Quality.}
Figure~\ref{fig:detail_comparison} shows examples of geometric detail that benefits from our prior. 
The first example shows that small features such as holes can be recovered from a very weak geometric signal in a noisy point cloud. Concavities, such as the space between the legs of the Armadillo, and fine shape details like the Armadillo's hand are also recovered more accurately in the presence of strong noise. In the heart example, the concavity makes it hard to estimate the correct normal direction based on only a local neighborhood, which causes SPR to produce artifacts. In contrast, the global information we use in our patches helps us estimate the correct sign, even if the local neighborhood is misleading.

\begin{figure}[t]
    \centering
    \includegraphics[width=\linewidth]{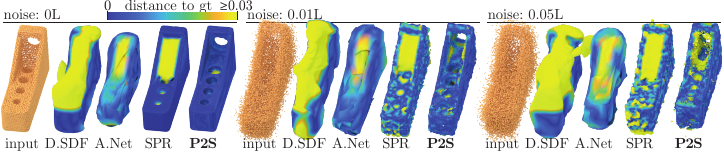}
    \caption{Effect of noise on our reconstruction. DeepSDF (D.SDF), AtlasNet (A.Net), SPR and Point2Surf (P2S) are applied to increasingly noisy point clouds. Our patch-based data-driven approach is more accurate than DeepSDF and AtlasNet, and can more robustly recover small holes and concavities than SPR.}
    \label{fig:noise}
\end{figure}

\paragraph{Effect of Noise.}
Examples of reconstructions from point clouds with increasing amounts of noise are shown in Figure~\ref{fig:noise}. 
Our learned prior for local patches and our coarse global surface information makes it easier to find small holes and large concavities. In the medium noise setting, we can recover the small holes and the large concavity of the surface. With maximum noise, it is very difficult to detect the small holes, but we can still recover the concavity.

\begin{figure}[t]
    \centering
    \includegraphics[width=\linewidth]{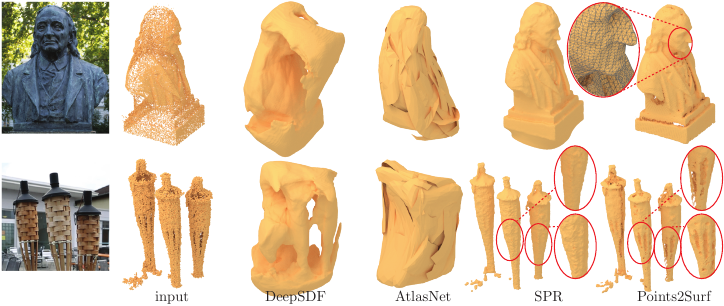}
    \caption{Reconstruction of real-world point clouds. Snapshots of the real-world objects are shown on the left. DeepSDF and AtlasNet do not generalize well, resulting in inaccurate reconstructions, while the smoothness prior of SPR results in loss of detail near concavities and holes. Our data-driven local prior better preserves these details.}
    \label{fig:real_world}
\end{figure}

\paragraph{Real-world data.}
The real-world point clouds in Figure~\ref{fig:teaser} bottom and Figure~\ref{fig:real_world} bottom both originate from a multi-view dataset~\cite{wolff2016noiseremoval} and were obtained with a plane-sweep algorithm~\cite{collins1996space} from multiple photographs of an object. We additionally remove outliers using the recent PointCleanNet~\cite{rakotosaona2019pointcleannet}. Figure~\ref{fig:real_world} top was obtained by the authors through an SfM approach. 
DeepSDF and AtlasNet do not generalize well to unseen shape categories. SPR performs significantly better but its smoothness prior tends to over-smooth shapes and close holes. Points2Surf better preserves holes and details, at the cost of a slight increase in topological noise. Our technique also generalizes to unseen point-cloud acquisition methods.

\subsection{Ablation Study}
\label{sec:ablation}
We evaluate several design choices in ($e_{\text{vanilla}}$) using an ablation study, as shown in Table~\ref{tab:ablation}. We evaluate the number of nearest neighbors $k=300$ that form the local patch by decreasing and increasing $k$ by a factor of $4$ ($k_{\text{small}}$ and $k_{\text{large}}$), effectively halving and doubling the size of the local patch. A large $k$ performs significantly worse because we lose local detail with a larger patch size. A small $k$ still works reasonably well, but gives a lower performance, especially with strong noise. We also test a fixed radius for the local patch, with three different sizes ($r_{\text{small}} := 0.05L$, $r_{\text{med}} := 0.1L$ and $r_{\text{large}} := 0.2L$).
A fixed patch size is less suitable than nearest neighbors when computing the distance at query points that are far away from the surface, giving a lower performance than the standard nearest neighbor setting.
The next variant is using a single shared encoder ($e_{\text{shared}}$) for both the global sub-sample $\mathbf{p}^s_x$ and the local patch $\mathbf{p}^d_x$, by concatenating the two before encoding them. The performance of $e_{\text{shared}}$ is competitive, but shows that using two separate encoders increases performance.
Omitting the QSTN ($e_{\text{no\_QSTN}}$) speeds-up the training by roughly 10\% and yields slightly better results. The reason is probably that our outputs are rotation-invariant in contrast to the normals of PCPNet.
Using a uniform global sub-sample in $e_{\text{uniform}}$ increases the quality over the distance-dependent sub-sampling in $e_{\text{vanilla}}$. This uniform sub-sample preserves more information about the far side of the object, which benefits the inside/outside classification.
Due to resource constraints, we trained all models in Table~\ref{tab:ablation} for 50 epochs only.
For applications where speed, memory and simplicity is important, we recommend using a shared encoder without the QSTN and with uniform sub-sampling.

\begin{table}[t]

\caption{Ablation Study. We compare \name ($e_{\text{vanilla}}$) to several variants and show the Chamfer distance relative to \name. Please see the text for details.}
\centering 
\small
\begin{tabular}{@{}rrrrrrrrrc@{}} 
& $r_{\text{small}}$ & $r_{\text{med}}$ & $r_{\text{large}}$ & $k_{\text{small}}$ & $k_{\text{large}}$ & $e_{\text{shared}}$ & $e_{\text{no\_QSTN}}$ & $e_{\text{uniform}}$ & $e_{\text{vanilla}}$ \\
\toprule
\ABC var-noise & 1.12 & 1.07 & 1.05 & 1.08 & 1.87 & 1.02 & 0.94 & 0.91 & \textbf{1.00} \\
\famous no-noise & 1.09 & 1.08 & 1.17 & 1.05 & 8.10 & 1.06 & 0.97 & 0.96 & \textbf{1.00} \\
\famous med-noise & 1.06 & 1.05 & 1.10 & 1.04 & 7.89 & 1.05 & 0.97 & 0.97 & \textbf{1.00} \\
\famous max-noise & 1.07 & 1.19 & 1.13 & 1.09 & 1.79 & 1.01 & 1.05 & 0.89 & \textbf{1.00} \\
\midrule
\textbf{average} & 1.08 & 1.11 & 1.11 & 1.07 & 4.14 & 1.03 & 0.99 & 0.92 & \textbf{1.00}
\end{tabular}

\label{tab:ablation}
\end{table}

\subsection{Timings}
\label{sec:timings}

We compare the wall-clock times of \name vanilla to the baselines. Because AtlasNet is very fast, we show the mean of 3 runs. We take the times of reconstructing the 100 shapes of the \ABC med-noise test set using 16 worker processes. With this, we can show a fair comparison that takes parallelization, loading times and different bottlenecks into account. We ran the timings on a consumer-grade PC with a Ryzen 7 3600X, 64 GB DDR4 RAM and a GTX 1070. The mean times per shape in seconds are: \name 712.1, DeepSDF 199.5, SPR 157.5, AtlasNet 0.1. Since DeepSDF and SPR need normals, we included 156.5 seconds for the normal estimation using PCPNet.

The heuristic and sign propagation allows us to reduce the number of inferred query points to 1.67\% of a full grid in \ABC var-noise. Assuming linear scaling, the inference time is reduced from almost 12 hours to 11.5 minutes. This speed-up comes at the cost of 19.2 seconds per shape. The heuristic is less efficient with noise. For the \ABC no-noise, the number of query points is reduced to 0.77\% and for \ABC max-noise to 2.33\%.

We trained a smaller version of $e_{\text{uniform}}$ with only about 15\% parameters. The mini version reduces the reconstruction time by roughly 68\% compared to $e_{\text{vanilla}}$ (from 11.5 to 3.7 minutes per mesh), at the cost of an increase in the reconstruction error of roughly 55\% on \ABC var-noise (Chamfer distance from 150.6 for $e_{\text{vanilla}}$ to 234.3 for the small version of $e_{\text{uniform}}$).

\section{Conclusion}

We have presented \name as a method for surface reconstruction from raw point clouds. Our method reliably captures both geometric and topological details, and generalizes to unseen shapes more robustly than current methods.

The distance-dependent global sub-sample may cause inconsistencies between the outputs of neighboring patches, which results in bumpy surfaces.

One interesting future direction would be to perform multi-scale reconstructions, where coarser levels provide consistency for finer levels, reducing the bumpiness.
This should also decrease the computation times significantly. Finally, it would be interesting to develop a differentiable version of Marching Cubes to jointly train SDF estimation and surface extraction.

\section{Acknowledgements}

This work has been supported by the FWF projects no. P24600, P27972 and P32418 and the ERC Starting Grant SmartGeometry (StG-2013-335373).

{\small
\bibliographystyle{splncs04}
\bibliography{points2surf}
}

\end{document}